\documentclass[sigconf]{acmart}
\usepackage{multirow}
\usepackage{balance}
\usepackage{color}
\AtBeginDocument{%
  \providecommand\BibTeX{{%
    \normalfont B\kern-0.5em{\scshape i\kern-0.25em b}\kern-0.8em\TeX}}}

\copyrightyear{2021} 
\acmYear{2021} 
\setcopyright{acmcopyright}\acmConference[MM '21]{Proceedings of the 29th ACM
International Conference on Multimedia}{October 20--24, 2021}{Virtual Event, China}
\acmBooktitle{Proceedings of the 29th ACM International Conference on Multimedia
(MM '21), October 20--24, 2021, Virtual Event, China}
\acmPrice{15.00}
\acmDOI{10.1145/3474085.3475636}
\acmISBN{978-1-4503-8651-7/21/10}

\begin{document}
\fancyhead{}
\fancyhf{}
\fancyhead[L]{ACM MM 2021 Oral Paper: Deep Learning for Multimedia}
\fancyhead[R]{Ning Wang and Guangming Zhu, et al.}
\title{Spatio-Temporal Interaction Graph Parsing Networks for Human-Object Interaction Recognition}

\author{Ning Wang* \quad Guangming Zhu*$^{\dagger}$ \quad Liang Zhang }
\author{ \quad Peiyi Shen \quad Hongsheng Li \quad Cong Hua}

\makeatletter
\def\authornotetext#1{
\if@ACM@anonymous\else
    \g@addto@macro\@authornotes{
    \stepcounter{footnote}\footnotetext{#1}}
\fi}
\makeatother
\authornotetext{Both authors contributed equally to this research.}
\authornotetext{Corresponding author.}

\affiliation{%
  \institution{School of Computer Science and Technology, Xidian University, Xi’an, Shaanxi, China}}
\email{{wangning2049,cong.huaa}@gmail.com, {gmzhu,liangzhang,pyshen}@xidian.edu.cn, hsli@stu.xidian.edu.cn}

\def\authors{Ning Wang, Guangming Zhu, Liang Zhang, Peiyi Shen, Hongsheng Li, Cong Hua}



\begin{abstract}
For a given video-based Human-Object Interaction scene, modeling the spatio-temporal relationship between humans and objects is the important cue to understand the contextual information presented in the video.
With the efficient spatio-temporal relationship modeling, it is possible not only to uncover contextual information in each frame, but to directly capture inter-frame dependencies as well.
Capturing the position changes of human and objects over the spatio-temporal dimension is more critical when significant changes in the appearance features may not occur over time.
When utilizing appearance features, the spatial location and the semantic information are also the key to improve the video-based Human-Object Interaction recognition performance.
In this paper, Spatio-Temporal Interaction Graph Parsing Networks (STIGPN) are constructed, which encode the videos with a graph composed of human and object nodes. 
These nodes are connected by two types of relations: (i) intra-frame relations: modeling the interactions between human and the interacted objects within each frame. (ii) inter-frame relations: capturing the long range dependencies between human and the interacted objects across frame.
With the graph, STIGPN learn spatio-temporal features directly from the whole video-based Human-Object Interaction scenes.
Multi-modal features and a multi-stream fusion strategy are used to enhance the reasoning capability of STIGPN.
Two Human-Object Interaction video datasets, including CAD-120 and Something-Else, are used to evaluate the proposed architectures, and the state-of-the-art performance demonstrates the superiority of STIGPN.
Code for STIGPN is available at \href{https://github.com/GuangmingZhu/STIGPN}{\textcolor{blue}{https://github.com/GuangmingZhu/STIGPN}}.
\end{abstract}



\begin{teaserfigure}
  \center
  \includegraphics[width=\textwidth]{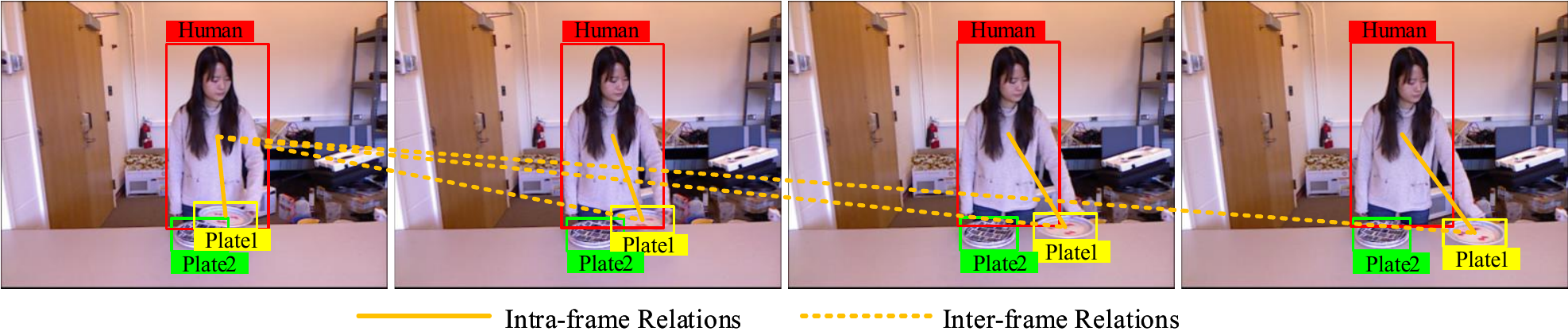}
  \caption{Two important cures exist to recognize Human-Object Interaction in video: First, model the global interaction relationship between human and the objects being interacted with. For clarity, only the inter-frame relations between the first frame and the rest of the frames are visualized.
Second, the spatial location trajectory and semantic labels can be crucial clues for recognizing when significant changes in the appearance features may not transpire during a Human-Object Interaction.}
  \label{fig:fig1}
\end{teaserfigure}

\maketitle

\section{Introduction}\label{section:1}
As one of the most fundamental tasks in scene understanding, Human-Object Interactions (HOI) has attracted significant attention in recent years. HOI detection methods, including video-based\cite{materzynska2020something,koppula2013learning,praneeth2020lighten} and image-based\cite{interactiveness,knowledgeGraph,intend,coco_dataset,hico}, have been widely studied. 
Generally, these methods visually understand the scene on the basis of existing mature methods, such as object detection\cite{hu2018relation,redmon2016you,erhan2014scalable,szegedy2014scalable}, action recognition\cite{zhu2020topology,yan2018spatial,yao2012action,si2019attention,zhang2020context}, visual relationship detection\cite{shang2017video,qian2019video,sun2019video,tsai2019video,shang2019annotating}, and so on. 
In this paper, we especially focus on the problem of learning HOI from videos. 
Video-based HOI aim to recognize the kind of interaction and which objects the human interacts with. 
For example, as shown in Fig. \ref{fig:fig1}, the goal is to recognize the HOI category: human moves the plate. 
Video-based HOI recognition requires greater understanding of spatio-temporal contextual information for better inference.

A straightforward method of video-based HOI recognition is to use deep learning end-to-end architectures, such as convolutional neural networks (CNN) and recurrent neural network (RNN), or 3D convolutional neural networks (3DCNN) \cite{tran2015learning} to learn the spatio-temporal features. However, all these architectures capture the spatio-temporal contextual information from the whole video scene instead of capturing the key instances.
Moreover, they fail to directly capture the spatio-temporal dependence between human and objects. 
Graphs are widely used to represent non-grid structures. 
When recognizing HOIs on the basis of human and object instances, graphs are the natural choice for modeling the spatio-temporal dependence. 
These methods usually detect humans and objects first, and then recognize HOIs from the spatio-temporal evolution of the graphs in which the nodes represent detected human and objects. 
Recently, many methods have attempted to model video-based HOI scenes on the basis of the graph structure.
Graph Parsing Neural Network (GPNN) \cite{qi2018learning} represents HOI structures with a single spatial graph and automatically parses the optimal graph structures in an end-to-end manner. GPNN can model a video in a single spatial graph, which benefits from the special spatio-temporal features as input features provided by the CAD-120 \cite{koppula2013learning} dataset. However, given the unavailability of these features, GPNN will find it difficult to perform equally well on other datasets.
The LIGHTEN \cite{praneeth2020lighten} models a video with graph sequences, which connects the instances of each other in each frame to build a graph. 
However, they must stack very deep GCNs and RNNs to capture spatio-temporal features from these graph sequences.

So is there a graph structure or method to capture spatio-temporal features more efficiently?
Let us reconsider the video HOI scene shown in Fig. \ref{fig:fig1}. 
How do humans effortlessly recognize the HOI category in videos? 
There could be two important clues that can solve this problem:
\textbf{First}, 
the spatio-temporal relationship between the human and the interacting objects is critical to our HOI recognition.
This means that attention should be paid not only to the objects being interacted with in the current frame, but also to the ones in other frames. 
As shown in the Fig. \ref{fig:fig1}, the status change of the plate being held deserve greater attention than the stationary plate and observing the state changes of the plate over time is beneficial for identifying HOI. 
Thus, the \textit{Intra-frame Relation} links need be established between the human and the interacting object in each frame and the \textit{Inter-frame Relation} links between the human and the interacting objects across frame.
This makes it possible not only to uncover contextual information in each frame but also to directly capture dependencies across time because of the complementarity of the intra-frame relation and the inter-frame relation.
These settings can capture local and global long-range dependencies by directly pairwise comparison of the features between human and object at all space-time location.
\textbf{Second}, capturing the appearance features of human and objects over the spatio-temporal dimension is crucial for video-based HOI recognition. 
However, significant changes may not emerge in the appearance features of human and objects over time when the views only demonstrate local and inapparent interactions. For example, significant changes in the appearance features of the human and the plate do not transpire, but the synchronous changes in the position of hands and the plate exactly indicate the HOI i.e., "human moves the plate." Therefore, utilizing the features beyond the appearance features, e.g., the spatial location and the semantic information of human and objects, become the keys to improve the interpretable HOI recognition.

On the basis of the above two clues, Spatio-Temporal Interaction Graph Parsing Networks (STIGPN) are constructed, which encode the videos with a graph composed of human and object nodes, and automatically learn the spatio-temporal relationship evolutions. First, multimodal features, including the appearance features, the spatial location, and the semantic information, are used for the feature representation of the input nodes of the graph. Then, a graph is built to represent the spatio-temporal relationship.
Nodes in the graph are connected by two types of edge: intra-frame relation and inter-frame relation. 
(i) Intra-frame relations model the interactions between the human and the interacted objects within each frame. (ii) Inter-frame relations capture the long range dependencies between the human and the interacted objects across frame.
Finally, the graph evolution operations are performed to extract the spatio-temporal features between graph nodes for the final HOI recognition.
Despite not using the groundtruth-based pre-computed features and the small amount of data available for training from videos, our networks achieve the state-of-the-art performance on the CAD-120 \cite{koppula2013learning} and Something-Else \cite{materzynska2020something} datasets. 


The main contributions of this paper can be summarized as follows: 
\textbf{First}, a novel and efficient spatio-temporal graph is proposed, which can directly model the global relationship between the human and the object to be interacted and capture the state change of the interacting objects across frames. \textbf{Second}, a framework which takes full use of visual, spatial and semantic features is proposed. \textbf{Third}, the state-of-the-art performances on two benchmark video-based HOI datasets are achieved.

\section{Related Work}
In this section, some existing technologies about human-object interaction detection and multi-streams neural networks are reviewed.

\textbf{Human-Object Interaction Detection.}
The HOI detection can be divided into two parts: 1) image-based HOI detection and 2) video-based HOI detection. The image-based HOI detection aims to recognize the interaction relationship between each pair of a human and an object from an image. This requires the model to utilize the contextual information in the scene. Georgia et.al.\cite{gkioxari2018detecting} proposed InteractNet that predicts the activity density at the location of the target object on the basis of the appearance of the detected human. Gao et al.\cite{gao2018ican} introduced an instance-centric attention-based network to selectively aggregate and recognize features related to HOIs. Recently, they were inspired by the design of two-stage object detectors and proposed dual relation graph in \cite{gao2020drg}, which can effectively capture the discriminative clues in the scene and solve the fuzzy problem of local prediction. Liang et al.\cite{liang2020visual} proposed a visual-segmantic graph attention network, which effectively aggregates contextual visual, spatial, and semantic information to eliminate the ambiguities of subsidiary relations. Furthermore, they contribute a Pose-based Modular Network (PMN) in \cite{liang2020pose} which explores the absolute pose features and relative spatial pose features to improve HOI detection. 

Unlike image-based HOI detection that received significant attention, video-based HOI detection is relatively less explored. This task is typically related to the prediction of human activity and objects affordance. Jain et al.\cite{jain2016structural} proposed a scalable method for casting an arbitrary spatiotemporal graph as a rich RNN mixture and for evaluating the task of HOI. Xiao et al.\cite{xiao2019reasoning} introduced a dual attention network model that weights the important features for objects and actions to reason about human-object interactions. Qi et al. \cite{qi2018learning} proposed the GPNN which represents HOI structures with graphs and automatically parses the optimal graph structures in an end-to-end manner. Sai et al. \cite{praneeth2020lighten} presented a hierarchical approach, named LIGHTEN, to effectively capture spatio-temporal cues. In this study, spatio-temporal graph parsing networks are constructed, which encode the videos with graph sequences and learn the spatio-temporal relationship evolutions over the temporal dimension and the spatial graph topologies.

\textbf{Multi-Stream Neural Network for Action Recognition.}
Recently, multi-stream neural networks have been successful for action recognition in videos. Singh et al.\cite{singh2016multi} contributed a multi-stream bi-directional recurrent neural network(BiRNN), which uses two different streams of information(i.e., motion and appearance) for action recognition. Feichtenhofer et al.\cite{feichtenhofer2016convolutional} presented a convolutional two-stream network and fused the appearance and motion information for video action recognition. Inspired by their work, Tu et al.\cite{tu2018multi} introduced a human-related multi-stream CNN architecture that integrates appearance, motion, and human-related regions. Zang et al.\cite{zang2018attention} proposed an Attention-based temporal weighted CNN, which uses three CNN streams to process spatial RGB images, temporal optical flow images, and temporal warped  optical flow images, respectively. Recently, Graph Convolutional Network (GCN) \cite{kipf2016semi} have extended CNNs to deal with non-grid structures and has achieved remarkable performance for action recognition. Shi et al.\cite{shi2020skeleton} contributed a multi-stream attention-enhanced adaptive graph convolutional neural network for skeleton-based action recognition. 

Generally, early and late feature fusions are widely used. Early fusion can enhance the input features, but the dimension of the imbalance of multimodal features may also suppress some features during training. Late fusion can overcome the shortcoming, but the separate inference can not utilize the mutual promotion of multimodal features. In this study, we first early fuse the visual and spatial features, the spatial and semantic features, respectively. Next, they are fed into the two-stream STIGPN for late fusion.

\begin{figure*}[htb]
\centering
\includegraphics[width=\linewidth]{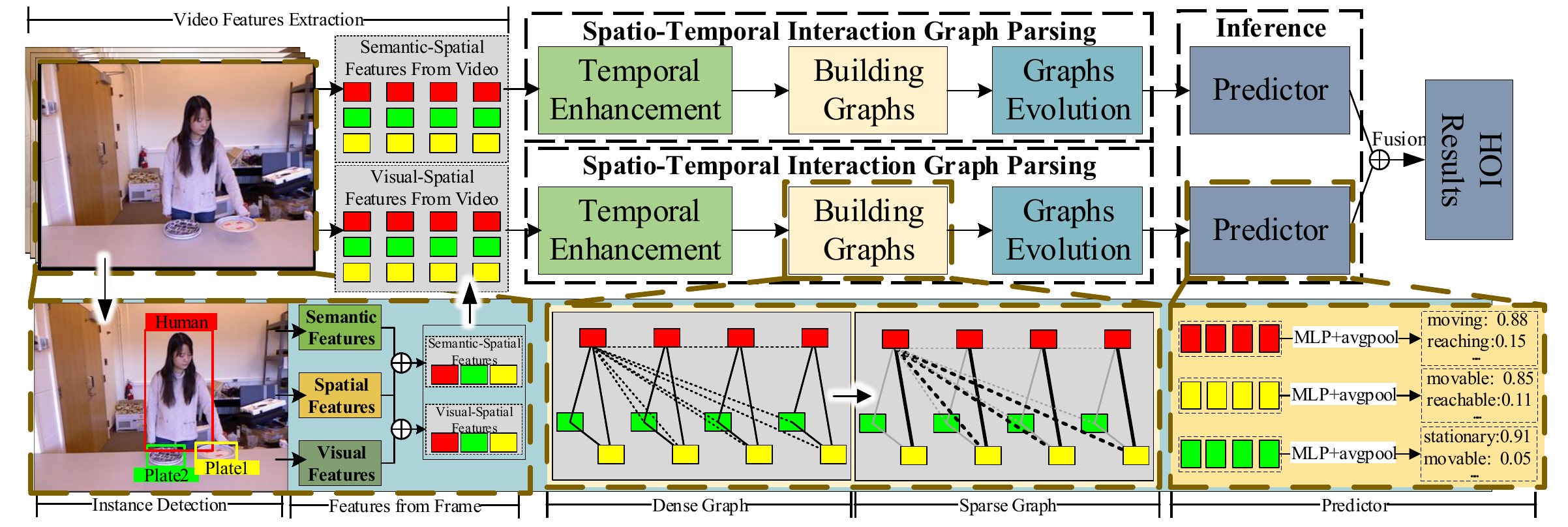}
\caption{Overview of the proposed networks. Our method first extracts two sets of the visual-spatial feature sequences and the spatial-semantic feature sequences from the input video. 
Then, they are fed into the two-stream networks to extract the spatio-temporal features respectively. 
Finally, the prediction result of each stream is obtained by using the prediction module and  fusion them to report the final HOI recognition results.
}
\label{fig:overall}
\end{figure*}

\section{Overview}
Our goal is to identify Human-Object Interactions in videos. 
Specifically, in videos, sub-activity labels (e.g., moving) for humans and a set of affordances labels (e.g., movable, stationary) for objects must be predicted. 
The overview of the proposed architecture is visualized in Fig. \ref{fig:overall}.
For the input video, an object detection networks, such as Faster R-CNN\cite{ren2016faster},is first used to locate a set of candidate instance on each video frame.
And the visual, spatial, and semantic features of each candidate instance are extracted independently.
In this study, the visual and spatial features are combined as visual-spatial features, and joint the spatial and semantic features as spatial-semantic features.
The two joint feature sequences extracted from the video are used as the input of our two-stream model.

Within each stream, we hope to build a graph that efficiently captures spatio-temporal features.
Generally, in a HOI scene, the state of the human and the interacting objects will change significantly.
Thus, a \textit{Temporal Enhancement} module is first performed on the joint feature sequences to allow the subsequent modules to focus on the salient parts of the feature.
Then, we constrcut a dense graph where the nodes correspond the human/objects detected from the video and parse a sparse graph from it.
The joint feature sequences are used to instantiate the corresponding nodes of the dense and the sparse graphs.
For simplicity, the sparse graph is decomposed into two subgraphs sub-graphs, which have the same nodes but different connections: the intra-frame relation graph and the inter-frame relation graph.
With the sparse graph, a \textit{Graphs Evolution} module is executed to extract the spatio-temporal features between graph nodes. 
Finally, always one \textit{Prediction} module exists for the final HOI classification. The prediction results of the two-streams are averaged to report the final performance.

\section{Proposed Approach}
In this section, the proposed STIGPN architecture is described in detail. 
Specifically, three components of STIGPN, i.e., \textbf{\textit{Video Feature Extraction}}, \textbf{\textit{Spatio-Temporal Interaction Graph Parsing}}, and \textbf{\textit{Inference}}, are described successively. The implementation details are provided in the end of this section.



\subsection{Video Feature Extraction}

Given a video $I$ with $T$ frames, $M$ instances of either human or object class, an object detection network, such as Faster R-CNN\cite{ren2016faster}, is first used to locate a set of candidate instance on each video frame. Each candidate instance comes with a bounding box and a category. Then, multi-object tracking is used to determine a correspondence between candidate instances in different video frames. 
On the basis of the above detected results, the visual, spatial, and semantic features of each candidate instance on video are extracted independently.

\textbf{Visual Features.} In this paper, the pre-trained ResNet-50 model is used to extract visual features due to limited data (the CAD-120 dataset has only 120 RGB-D videos). Given the human/objects bounding boxes, the first step is to extract the ROI crops from the original images and reshape them to the input dimension of ResNet-50. Next, these ROIs are fed into ResNet-50 and the outputs of the last fully-connected layer are extracted as the visual features of human or object instances.

\textbf{Spatial Features.}
Spatial features represent the relationship information between candidate human/objects. In this paper, a simple yet efficient representation of spatial features is used. Given a candidate human/object bounding box with its width $W$ and height $H$, a quadruple is defined as $[x_c,y_c,W,H]$ where $(x_c,y_c)$ is the center position of the bounding box. A fully connected layer is used to transform the quadruples to $d$-dimensional spatial features.

\textbf{Semantic Features.}
Semantic features represent the identity information of human and objects. In this paper, a learnable $d$-dimensional embedding is trained to represent the semantic features of human/objects. This can help our model be applicable to more datasets because a general method is used to embed all kinds of objects.

The combination of the above three kinds of features can be used as the instance feature representation. 
In this study, the visual and spatial features are combined as the visual-spatial features, and the spatial and semantic features are combined as spatial-semantic features.
Then, they were fed into the two-stream network for HOI learning respectively.

\subsection{Spatio-Temporal Interaction Graph Parsing}
The input can be defined as $\{X_m,c_m\}_{m=1,..,M}$ with a class label $c_m$ where $M$ is the number of instances (either human or object) and $X_m = \{x^t_m\}_{t=1,..,T}$ denote the temporal sequence of frame-level features.
On the basis of the inputs, the Spatio-Temporal Interaction Graph Parsing module is proposed to extract high-level features for the final recognition.

\textbf{Temporal Enhancement.}
The changing orientations of human/objects are captured to help to enhance the extraction of the salient human-object interaction pairs. Therefore, a BiRNN is used to evolve the temporal relationship over the recurrent steps. 
The BiRNN process can be expressed as 
\begin{equation}
\overrightarrow{h}_{m}^t=f(\overrightarrow{W}_x x_{m}^t + \overrightarrow{W}_h \overrightarrow{h}_{m}^{t-1} + \overrightarrow{b}_h),
\end{equation}
\begin{equation}
\overleftarrow{h}_{m}^t=f(\overleftarrow{W}_x x_{m}^t + \overleftarrow{W}_h \overleftarrow{h}_{m}^{t-1} + \overleftarrow{b}_h),
\end{equation}
\begin{equation}
y_{m}^t=g(W_y [\overrightarrow{h}_{m}^t,\overleftarrow{h}_{m}^t] + b_y),
\end{equation}
\noindent where $\overrightarrow{W}_x$ and $\overleftarrow{W}_x$ are the weight matrices that map the input to the hidden state, and $\overrightarrow{W}_h$ and $\overleftarrow{W}_h$ are the transition matrices between the hidden states in two adjacent time steps. $W_y$ is the the weight matrix that maps the hidden states to the output. $\overrightarrow{b}_h$, $\overleftarrow{b}_h$ and $b_y$ are the biases. $f$ and $g$ are activation functions. $[\cdot]$ is the concatenation operation.

In the current stage, BiRNN is implemented on the temporal sequence of frame-level features, without fusion of the features of different categories of instances. The instances that have significant changes in appearance or spatial features can be enhanced to some degree. Long Short-Term Memory (LSTM) network is not used, given that RNN is easier to train and has comparable performance when the sequences are not longly. The size of the used video datasets used in the experiments also restricts the network complexity. 


\textbf{Building Graphs.}
In this paper, a graph is constructed to represents the attribute of human and objects and the relationship between them. 
A dense graph is first constructed and then a sparse graph is parsed from it.
Formally, the dense graph with a size of $T\cdot M \times T\cdot M$ is defined as $G = (V,E)$ where $V$ is the set of nodes and $E$ is the set of edges.
Let $v_m^t \in V$ be the $m$-th node of the $t$-th graph, and the temporal enhanced features $y_{m}^t$ are used to instantiate the corresponding nodes $v_m^t$ of the graph.
The edge weights are initialized to be 1 for human-object edges and 0 for the rest. The adjacency matrix is dynamically learned via backpropagation.
To obtain richer structure information and enlarge the number of propagation neighborhoods during graph evolution, the graph is built in a bidirectional manner.

Graph structures reflect human-object interaction relationships. 
However, in the dense graph, the number of edges is close to the maximal number of edges.
Thus, a parsed graph is inferred, in which humans only have salient connections with the objects they are interacting with.
In this paper, the self-attention mechanism is used to reset the edge weights of the dense graph to infer a parsed graph.
Formally, let $y_{v_i^t}$ and $y_{v_j^t}$ denote the features of the source and the target nodes, the affinity between them can be computed as
\begin{equation}
Att(y_{v_i^t} , y_{v_j^t}) = LeakyReLU(\textbf{a}[W y_{v_i^t},W y_{v_j^t}])
\end{equation}
\noindent where $W$ is a shared weight matrix for node-wise feature transformation, and $\textbf{a}$ is a learnable weight vector for computing attention coefficients. The LeakyReLU activation is used to maintain weak connections. $[\cdot]$ is the concatenation operation.
For simplicity, the constructed large graph is decomposed into two subgraphs sub-graphs, which have the same nodes but different connections: the intra-frame relation graph and the inter-frame relation graph.

\textit{Intra-frame Relation Graph.}
The intra-frame relation graph reflect human-object interaction relationships in each frame. 
Specifically, a highly credible edge between the pair of nodes represented by the person and the interacting objects in each frame need be established.
Similar to \cite{yan2018spatial}, a set of nodes connected to $v_m^t$ can be represented as
\begin{equation}
B(v_m^t) = \{v_j^q | d(v_j^q,v_m^t) \leq 1,q=t \}
\end{equation}
where $d(v_j^q,v_m^t)$ denotes the minimum length of any path from $v_j^q$ to $v_m^t$.
For each node $v_m^t$, the affinity between the node $v_m^t$ and the node $v_j^q \in B(v_m^t)$ can be calculated by $Att(y_{v_m^t} , y_{v_j^q})$.
After calculating the affinity value of all node pairs in the graph, an affinity matrix with the same size of the adjacency matrix of the dense graph is obtained.
It is worth noting that for those node pairs that have no connected edges of the dense graph, their affinity value is directly set to 0.
Finally, normalization is performed on each row of the affinity matrix to obtain the adjacency matrix $A^{intra}$ of the intra-frame relation graph. 
The above steps of obtaining $A^{intra}$ can be more intuitively expressed as
\begin{equation}
A^{intra}=
\begin{cases}
Att(y_{v_m^t} , y_{v_j^t}) & v_j^t \in B(v_m^t) \\ 0 & other 
\end{cases}
\end{equation}

\textit{Inter-frame Relation Graph.}
Although the intra-frame relation graph captures the spatial dependence between the human and the interacting objects, it does not capture their long-term dependence and state correlations.
Therefore, the inter-frame relation graph is proposed, which is complementary to the intra-frame relation graph.
With the inter-frame relation graph, the correlations between the human and the interacting objects across frames can be identified. 
The process of parsing the inter-frame relation graph from the dense graph is similar to parsing the intra-frame relation graph.
Given a node $v_m^t$, a set of nodes connected to $v_m^t$ can be represented as
\begin{equation}
B'(v_m^t) = \{v_j^q | d(v_j^q,v_m^t) \leq 1,q\neq t \}
\end{equation}
$B'(v_m^t)$ represents a set of nodes, and the node in the set are on different frames with node $v_m^t$.
After the same operation as obtaining the intra-frame relation graph, the adjacency matrix $A^{inter}$ of the inter-frame relation graph is finally obtained.
Intuitively, $A^{inter}$ can be obtained by
\begin{equation}
A^{inter}=
\begin{cases}
Att(y_{v_m^t} , y_{v_j^q}) & v_j^q \in B'(v_m^t) \\ 0 & other 
\end{cases}
\end{equation}

\textbf{Graphs Evolution}
To perform reasoning on the graph, two methods are applied to update node features: graph neural networks (GCNs) and temporal fusion. 

\textit{Graph neural networks.} 
GCNs propagate messages via the connection relationship between the graph nodes, which dynamically updates the node feature vector. 
In this paper, the GCNs proposed in \cite{kipf2016semi} are used to evolve the intra-frame relation graph and the inter-frame relation graph, respectively.
To combine multiple graphs in GCNs, their output is concatenated. The above operation can be expressed as
\begin{equation}
Z^{st} = [A^{intra}YW^{intra},A^{inter}YW^{inter}]
\end{equation}
\noindent where $Y=\{y_m^t\}_{m=1,..,M}^{t=1,..,T}$ denote the input node features, $W^{intra}$ and $W^{inter}$ are the learnable weight matrix. $[\cdot]$ is the concatenation operation.

\textit{Temporal Fusion.}
With the intra-frame relation graph and the inter-frame relation graph, the GCNs module only propagate messages between people and objects, but not between homogeneous node pairs (including human-human, object-object).
Therefore, after the GCNs module, a BiRNN is performed on the nodes represented by instances of the same entity throughout time to capture the temporal dependencies.
The output $Z=\{z_{v_m^t}\}_{m=1,..,M}^{t=1,..,T}$ as the final video-level representation can be used to predict the HOI results.

\subsection{Inference}
Generally, always one prediction module exists for the final HOI classification. Furthermore, visual and semantic modalities have their respective superiorities for HOI understanding. Therefore, a two-stream strategy is employed to improve the network performance.  

\textbf{Prediction.}
The parsed graphs still contain human and object nodes. The final features of human nodes can be used to predict interaction categories, and the final features of object nodes can be used to predict an object's affordance labels (if applicable). When the dataset, such as CAD-120, has object affordance labels associated with HOI activities, the objects that the human is interacting with predicts the affordance labels, and other nodes predict the "Stationary" label to indicate the objects that are not involved in the HOI activities.

Given the final node features of the parsed graphs as $[z_{v_m^1},z_{v_m^2},...\\,z_{v_m^T}]$, a human-classifier $f_{readout}^h$ and an object-classifier $f_{readout}^o$ are executed on the human and object node features, respectively, for the final recognition. The prediction of interaction categories can be fused as 
\begin{equation}
p^h = \sum_{t=1}^Tf_{readout}^h(z_{h^t}) / T,
\end{equation}
\noindent where $z_h^t$ is the human node feature in the $t$-th frame. The $v_m$-th object's affordance label can be predicted as
\begin{equation}
p_{v_m}^o = \sum_{t=1}^Tf_{readout}^o(z_{{v_m^t}}) / T.
\end{equation}

\textbf{Two-Stream Fusion.}
A straightforward way to fuse the appearance, spatial and semantic features is to feed the concatenation of the three types of features into the STIGPN network. An alternative way is to construct a two-stream network to improve the performance. 

\textit{Visual-STIGPN.} In this stream, the appearance features of each node are augmented by concatenating the spatial features. This will enhance the feature representation ability of nodes by embedding the spatial location information.

\textit{Semantic-STIGPN.} Human can recognize HOI activities only by using the object category and information about the object's changing  spatial location. Therefore, the spatial and the semantic features are concatenated to feed them into the Semantic-STIGPN stream for the recognition.

The Visual-STIGPN and Semantic-STIGPN has the same network architecture, but different input node features. The prediction results of the two-streams are averaged to report the final performance.

\textbf{Loss Functions.}
We subject both streams' classifiers to standard Cross-Entropy losses $\mathcal{L}_v$ and $\mathcal{L}_s$, and each comprises human activity loss $\mathcal{L}_v^h$ (or $\mathcal{L}_s^h$) and object affordance loss (if applicable) $\mathcal{L}_v^o$ (or $\mathcal{L}_s^o$). The overall loss can be written as
\begin{equation}
\mathcal{L} = \mathcal{L}_v+\mathcal{L}_s=(\mathcal{L}_v^h+\lambda\mathcal{L}_v^o)+(\mathcal{L}_s^h+ \lambda\mathcal{L}_s^o).
\end{equation}

\subsection{Implementation Details}
In this section, the implementation details are discussed from two different point of views: model and training.

\textbf{Model.} Each video segment or sequence is uniformly sampled to a fixed frame number $T$ ($T=10$ for CAD-120 and $T=16$ for Something-Else). The ROI crops are extracted from each frame and are resized to a fixed size of $224\times 224\times 3$ for ResNet-50. A linear layer is used to transform the 2048-dimensional ResNet-50 feature vectors to 1024-dimensional visual feature vectors. 
A fully connected layer is used to transform the quadruples to $d$-dimensional spatial features. 
with architecture "Linear128-BatchNorm-ReLU-Linear256-BatchNorm-ReLU" is used to transform the spatial quadruples into 256-dimensional feature vectors. The embedded semantic features are 128-dimensional vectors. The BiRNN in the Temporal Enhancement module keeps the feature dimension unchanged, the Spatial-Temporal Interaction Reasoning Network module transforms the node features into 1024-dimension, and the BiRNN in the Temporal Fusion module outputs 2048-dimensional features. The last MLP classifiers have the architecture of "Linear2048-BatchNorm-ReLU-Linear512-BatchNorm-ReLU-LinearC" where C is the category number.

\textbf{Training.} PyTorch framework and the DGL library\cite{wang2019deep} are used to implement our method. During training, Adam \cite{adam} optimizer with the initial learning rate of $2\times 10^{-5}$ is used. The learning rate decreases by 0.8 every 10 epochs. A total of 300 epochs are implemented on Nvidia TITAN X Pascal GPU. When training on CAD-120, both the sub-activity labels and the object affordance labels are used to supervise the prediction of all the nodes. While training on Something-Else dataset, only the activity labels are used to supervise the human node prediction.

\section{Experiments}

\subsection{Dataset}
Two benchmark datasets, i.e., CAD-120 \cite{koppula2013learning} and Something-Else \cite{materzynska2020something} are used to evaluate our proposed framework. 

\textbf{CAD-120:} The CAD-120 dataset is a video dataset, which has 120 RGB-D videos of four subjects performing ten daily indoor activities \emph{(e.g., microwaving food, taking food)}. Each activity is a long sequence of video, consisting of several segments which contain a finer-level sub-activity \emph{(e.g., the activity 'microwaving food' is divided into six sub-activities in a chronological order: 'reaching microwave', 'opening microwave', 'reaching food', 'moving food', 'placing food', 'closing microwave')}. 
In each video, the human is annotated with an sub-activity label from a set of ten sub-activity classes \emph{(e.g., moving, opening)}. Each object is annotated with an affordance label from a set of twelve affordance classes \emph{(e.g., movable, openable).}

The dataset provides multimodal data, but only the RGB images and the 2D bounding boxes annotation of human and objects are used in this paper. The sub-activity F1-score and object affordance F1-score are computed as the metric to evaluate the proposed method on HOI sub-activity recognition and anticipation. The dataset involves complex interactions, and thus human may be interacting with multiple objects during an activity. Therefore, it is necessary to learn the salient interaction relationships from the whole scene.

\textbf{Something-Else:} The Something-Else dataset is built on the Something-Something V2 dataset \cite{goyal2017something}, which has 112,795 videos of 174 categories of activities. The dataset provides a 2D bounding box of the hand (hands) and objects involved in an activity for each video frame. Particularly, the dataset forces the same combinations of activities and objects existed in the training set be absent in the testing set.

\subsection{Ablation Study}
The proposed STIGPN consists of three components: temporal enhancement, graph building and graph evolution. A BiRNN layer is used in the temporal enhancement component, while GCN and another BiRNN layer in the graph evolution. To verify the necessity of extracting the changing in  the status of human/objects and fusing the spatial and semantic features, some ablation studies are implemented.



\begin{table*}
	\caption{Ablation experiments of the impact of design choices on sub-activity and object affordance recognition of CAD-120 dataset.}
	\label{tab:ablation}
	\begin{tabular}{|c|c|c|c|}
		\hline
		Stream & Network Architecture  & \multicolumn{1}{c|}{\begin{tabular}[c]{@{}c@{}} Human \\ Sub-activity\end{tabular}} & \multicolumn{1}{c|}{\begin{tabular}[c]{@{}c@{}}Object \\ Affordance\end{tabular}} \\ \hline
		\multirow{4}{*}{\begin{tabular}[c]{@{}c@{}}One-Stream\end{tabular}} & Spatial GCN+BiRNN &79.5 &83.7 \\ 
		& Intra-frame Relation GCN+BiRNN &82.8  &83.1 \\ 
		& Inter-frame Relation GCN+BiRNN &83.5  &86.6 \\ 
		& Intra-frame Relation GCN+Inter-frame Relation GCN+BiRNN &86.1  &87.9 \\ 
		&\begin{tabular}[c]{@{}c@{}}BiRNN+Intra-frame Relation GCN+Inter-frame Relation GCN+BiRNN\end{tabular}&88.1 &90.4\\ \hline
		\multirow{1}{*}{Two-Stream} & \textbf{STIGPN(BiRNN+Intra-frame Relation GCN+Inter-frame Relation GCN+BiRNN)} &\textbf{91.9} &\textbf{92.0}\\ \hline
	\end{tabular}
\end{table*}

\textbf{Role of Temporal Enhancement.}
As claimed ahead, change means actions. To capture the changing orientations of human/objects ahead is helpful for HOI recognition. A variant of STIGPN, in which the temporal enhancement component is removed, is evaluated. 
This may make it difficult for the subsequent modules to focus on the salient parts of graphs.
The performance comparison between rows 4 and 5 in Table \ref{tab:ablation} demonstrates the necessity of temporal enhancement. The wider gap on the sub-activity F1-score compared with the affordance F1-score, also proves the importance of extracting status change for activity recognition.

\textbf{Role of Graph Structure.}
Traditional architecture, such as the GCN+RNN, learns the spatial features first and then fuses them along the temporal dimension. 
As shown in the baseline results in row 1 of Table \ref{tab:ablation},it performs poorly when applying the spatial GCNs (applying the GCNs on the constructed dense graph).
After that, as claimed ahead, a HOI scene may contain multiple objects, and only one or part of them are involved in the HOI process. The HOI recognition methods based on the detected object instances, must judge which object(s) the human is interacting with for high precision HOI recognition. Therefore, it is necessary to evolve the spatial graph topologies. 
With the intra-frame relation graph, a 3.3\% improvements on the sub-activity F1 score can be obtained.
Similarly, the clue that the state of an object changes with respect to human is helpful for HOI recognition.
Our method improves the sub-activity and object affordance F1 score by 4.0\% and 2.9\% again by building the inter-frame relation graph.
As shown in the row 4 of Table \ref{tab:ablation}, our method achieves significant gains in sub-activity, and object affordance F1 score benefited from combining two graphs together.
Furthermore, the input node features are already deep, thus learning the deeper features is no longer necessary. The changing components and their changing styles are more important, and learning them for short-term HOI processes is not difficult.

\textbf{Role of Two-Stream Fusion.}
To measure the multimodal features fusion and multi-stream fusion strategy, an experiment is designed, as shown in row 6 of Table 1, which fuses the appearance and spatial features, the spatial and semantic features, respectively, and feeds them into two-stream STIGPN. Compared with the experiment shown in row 5 of Table 1, the concatenation of the appearance, spatial and semantic features is fed into one-stream, and the two-stream STIGPN achieves a better performance than one-stream STIGPN. This improvement can be attributed to the multimodal features fusion and multi-stream fusion strategy, which utilizes the mutual promotion of multimodal features.
The full model achieve state-ofthe-art results on two challenging datasets, as illustrated in Tables \ref{tab:cad120_recog}-\ref{tab:something}.
%

\subsection{Comparison with State-of-the-arts}
\textbf{Comparison on CAD-120.} 
CAD-120 contains long activity video sequences, which comprise segments of sub-activities. HOI recognition is performed to predict sub-activities and object affordance labels of segments. Moreover, HOI anticipation can also be performed to predict the next segment's labels on the basis of the current segment data.
F1-score is used as the evaluation metric.
Tables \ref{tab:cad120_recog} and \ref{tab:cad120_antici} report the results for HOI recognition and anticipation, respectively. The proposed method outperforms the existing methods, such as ATCRF\cite{koppula2015anticipating},  Structural-RNN (S-RNN)\cite{jain2016structural}, GPNN\cite{qi2018learning}, and LIGHTEN\cite{praneeth2020lighten}. The proposed method achieves superior accuracy than the latest LIGHTEN method. 
The contextual information of inter-segment can also be used to improve the recognition accuracy. 
Therefore, Sai et al. \cite{praneeth2020lighten} proposed a segment-level temporal subnet (Seg-RNN) for segment-level fusion. 
However, applying this method may be difficult in real-time scene, hence this module is not included in our model. 
It is worth noting that our method does not perform segment-level fusion but is also outperforming the LIGHTEN with Seg-RNN. This exactly demonstrates the superiority of the proposed method.

\begin{figure}
	\centering
	\includegraphics[width=0.9\linewidth]{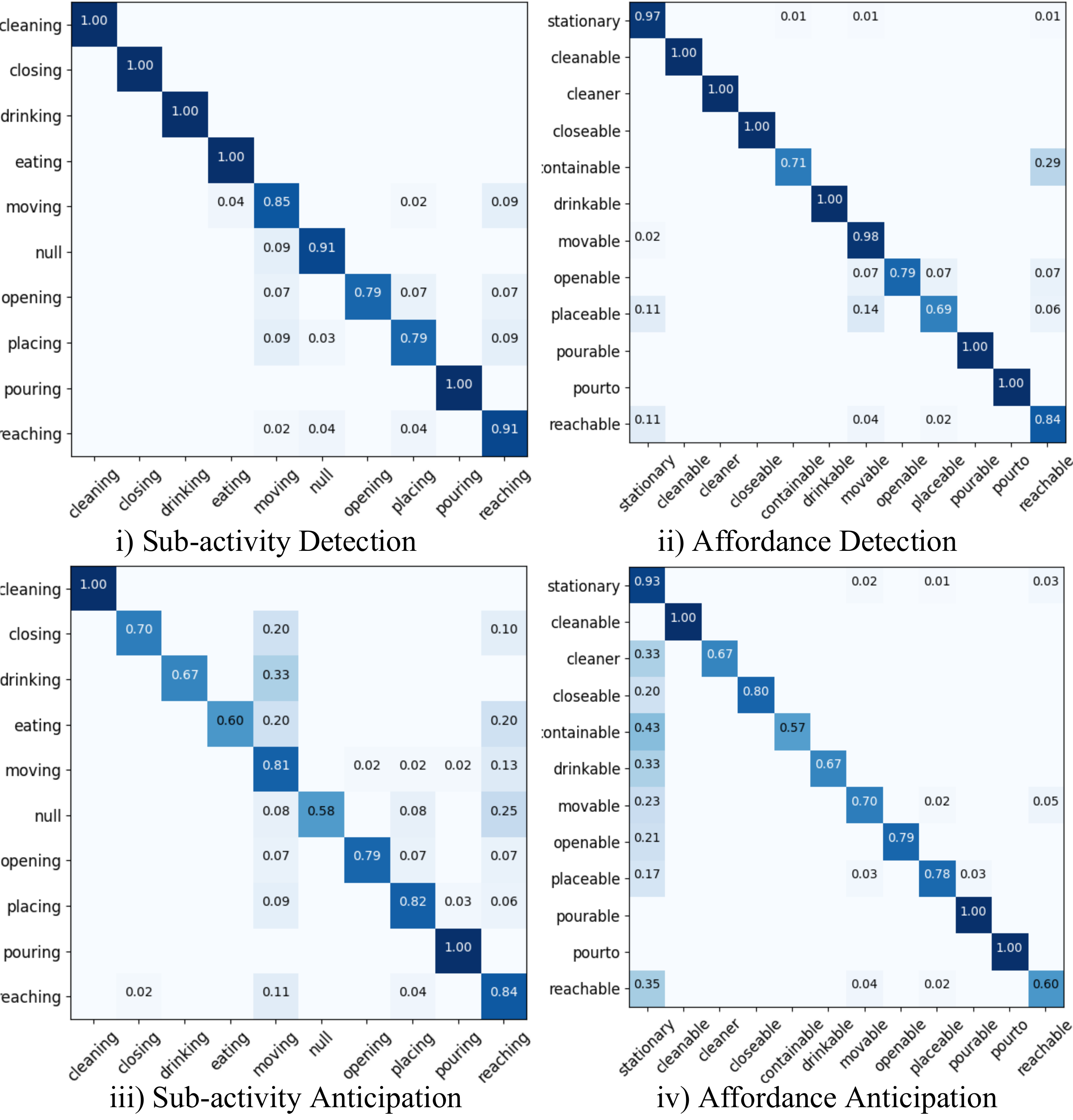}
	\caption{The confusion matrices for HOI recognition and anticipation on CAD-120 dataset.}
	\label{fig_cm}
\end{figure}

The confusion matrices for both HOI recognition and anticipation tasks on CAD-120 are displayed in Fig.\ref{fig_cm}. The affordance "containable" is easily misrecognized as "reachable", this is because when "placing" something in the microwave, it also involve similar tiny movements, such as "reaching". 
Our method performs well in other interaction scene with multiple objects, e.g., "cleaning something" or "pouring something".

\begin{table}[!htb]
	\centering
	\caption{A comparison of our approach with the existing methods on HOI recognition on CAD-120 dataset}
	\begin{tabular}{|c|c|c|c|c|}
		\hline
		\multirow{2}{*}{Method} & \multicolumn{2}{c|}{F1 Score in \%}\\ \cline{2-3}
		&Sub-activiy(\%)&Affordence(\%)\\ \hline
		ARCRF\cite{koppula2015anticipating}  & 80.4 & 81.5\\
		S-RNN\cite{jain2016structural}  & 83.2 & 88.7\\
		S-RNN (multi-task)\cite{jain2016structural}  & 82.4 & 91.1\\
		GPNN\cite{qi2018learning}  & 88.9 & 88.8\\
		LIGHTEN w/o Seg-RNN\cite{praneeth2020lighten} & 85.9 & 88.9\\
		LIGHTEN\cite{praneeth2020lighten} & 88.9 & \textbf{92.6}\\  \hline
		\textbf{STIGPN}                                       & \textbf{91.9} & 92.0\\  \hline
	\end{tabular}
	\label{tab:cad120_recog}%
\end{table}

\begin{table}[!htb]
	\centering
	\caption{A comparison of our approach with the existing methods on HOI anticipation on CAD-120 dataset.}
	\begin{tabular}{|c|c|c|c|c|}
		\hline
		\multirow{2}{*}{Method} & \multicolumn{2}{c|}{F1 Score in \%}\\ \cline{2-3}
		&Sub-activiy(\%)&Affordence(\%)\\ \hline
		ARCRF\cite{koppula2015anticipating}  & 37.9 & 36.7\\
		S-RNN\cite{jain2016structural}  & 62.3 & 80.7\\
		S-RNN (multi-task)\cite{jain2016structural}  & 65.6 & 80.9\\
		GPNN\cite{qi2018learning}  & 75.6 & \textbf{81.9}\\
		LIGHTEN w/o Seg-RNN\cite{praneeth2020lighten}  & 73.2 & 77.6 \\
		LIGHTEN\cite{praneeth2020lighten}  & 76.4 & 78.8\\ \hline
		\textbf{STIGPN}  & \textbf{81.1} & 81.8\\ \hline
	\end{tabular}
	\label{tab:cad120_antici}%
\end{table}

\begin{table}[!htb]
	\centering
	\caption{A comparison of our approach with the existing methods on HOI recognition on Something-Else dataset.}
	\begin{tabular}{|c|c|c|}
		\hline
		Method & Top-1(\%) & Top-5(\%)\\ \hline
		STIN+OIE\cite{materzynska2020something}  & 51.3 & 79.3\\
		STIN+OIE+NL\cite{materzynska2020something}  & 51.4 & 79.3\\ 
		STGCN\cite{shi2020skeleton}  & 54.4 & 81.4\\ \hline
		I3D\cite{wang2018videos}  & 46.8 & 72.2\\
		STRG\cite{wang2018videos} & 52.3 & 78.3 \\
		I3D+STIN+OIE+NL\cite{materzynska2020something}  & 54.6 & 79.4\\ 
		I3D,STIN+OIE+NL\cite{materzynska2020something}  & 58.1 & 83.2\\ \hline
		\textbf{STIGPN}  & \textbf{60.8} & \textbf{85.6}\\ \hline
	\end{tabular}%
	\label{tab:something}%
\end{table}%

\textbf{Comparison on Something-Else.} 
We further evaluate the proposed method on the Something-Else dataset, and the comparison results are reported in Table.\ref{tab:something}.
The proposed method outperforms the state-of-the-art methods, including STGCN \cite{shi2020skeleton} which has nine layers of adaptive spatial graph convolution and temporal convolution. 
In \cite{materzynska2020something}, the appearance features are additionally used. I3D is widely used for video-based action recognition. The I3D using ResNet-50 backbone \cite{wang2018videos} is employed for comparison. "STRG" is short for Space-Time Region Graph \cite{wang2018videos}. "I3D+STIN+OIE+NL" means combining the appearance features from the I3D model and the features from the STIN+OIE+NL model through joint learning. "I3D,STIN+OIE+NL" means a simple ensemble model combining the separately trained I3D and the trained STIN+OIE+NL model. 
The comparison results in Table \ref{tab:something} show that the proposed method outperforms all the above methods. 
Ultimately, the proposed methods achieve the state-of-the-art performance, and still gains the top-1 and top-5 accuracy by 2.7\% and 2.4\%, respectively, even if our method only uses the 2D appearance features of ResNet-50 but not the 3D features of I3D.

\subsection{Visualization of Parsed Graphs}
\begin{figure}
	\centering	
	\includegraphics[width=0.8\linewidth]{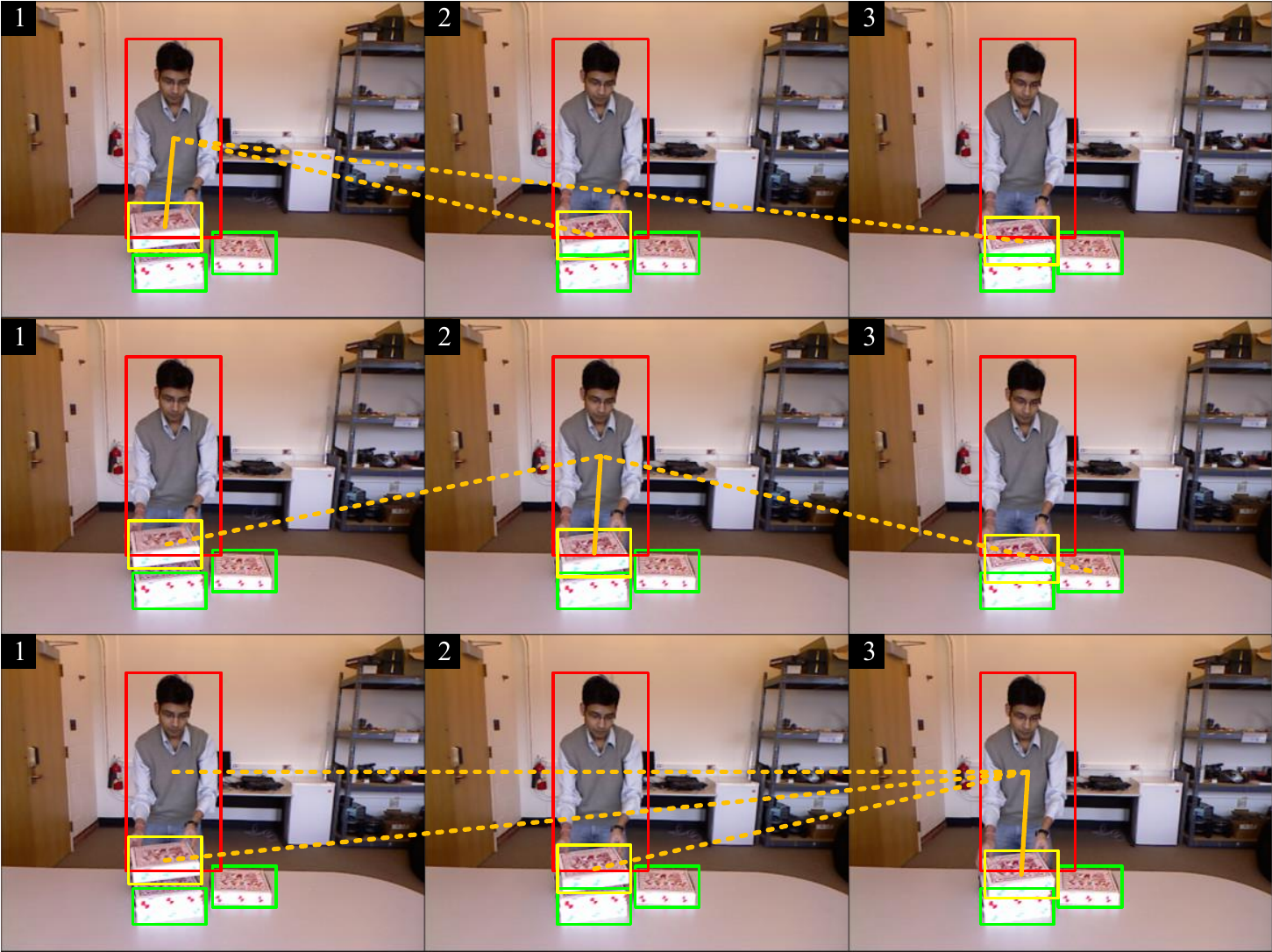}
	\caption{Visualization of the parsed graphs of Visual-STIGPN extracted from a video segment CAD-120 dataset. The solid lines indicate the interaction relationships between human and objects in space. The dotted lines indicate the interaction relationships between human and objects across time.}
	\label{fig_vis}
\end{figure}

When constructing the dense graphs, the edges between human and objects are uniformly set to 1. 
The parsed graphs are visualized to check whether STIGPN can find the object(s) the human is interacting with. 
For simplicity, the two parsed graph topologies of the Visual-STIGPN are visualized. 
For the intra-frame relation graph, an edge in the edge set $\{B(v_h^t) \rightarrow v_h^t\}$ is taken with the largest affinity value for visualization where $v_h^t$ is the node corresponding to the human in frame $t$.
For the inter-frame relation graph, some edges in the edge set $\{B'(v_h^t) \rightarrow v_h^t\}$ corresponding to the top n affinity value are used for visualization.
Fig.\ref{fig_vis} provides some visualization examples. It can be seen that the edges between the human and the object being interacted with have much larger values.

\section{Conclusion}
In this paper, a novel video-based HOI recognition architecture is presented, i.e., the Spatio-Temporal Interaction Graph Parsing Networks. 
Our framework is composed of three steps, namely, Video Feature Extraction, Spatio-Temporal Interaction Graph Parsing, and Inference.
The visual, spatial and semantic features are utilized by combining early and late fusion strategy. 
The proposed method aims to learn the spatio-temporal relationship evolution and find the objects involved in HOI process from the background objects on the basis of the parsed graphs. 
The visualization of the parsed graphs demonstrates that the proposed architecture can extract the salient human-object interaction pairs effectively. 
Experimental results on two publicly available benchmark video HOI datasets show that our architecture outperforms state-of-the-art methods. 

\begin{acks}
This work is partially supported by the National Natural Science Foundation of China under Grant No.62073252. and No.62072358.,
and the National Key R\&D Program of China under Grant No. 2020YFF0304900.

\end{acks}

\newpage

\bibliographystyle{ACM-Reference-Format}
\balance
\bibliography{sample-base}

%
%
%
%
%
%
%
%

\end{document}